\documentclass[smallextended]{svjour3}

\usepackage{booktabs} 
\usepackage{graphicx}
\usepackage{subfigure}
\usepackage{amssymb,amsmath}
\usepackage[ruled,linesnumbered]{algorithm2e}
\begin{document}

\title{Generative Temporal Link Prediction via Self-tokenized Sequence Modeling}

\author{Yue Wang$^{1,4}$ \and Chenwei Zhang$^2$ \and Shen Wang$^2$ \and Philip S. Yu$^2$ \and Lu Bai$^1$\footnote{Lu Bai is the corresponding author, \email{bailucs@cufe.edu.cn}} \and Lixin Cui$^1$ \and Guandong Xu$^3$ 
}

\institute{1. Central University of Finance and Economics, Beijing, P.R. China.\\
2. Department of Computer Science, University of Illinois at Chicago, Chicago, USA.\\
3. Advanced Analytics Institute, University of Technology, Sydney, Australia.\\
4. State Key Laboratory of Cognitive Intelligence, iFLYTEK, P.R. China
}

\maketitle

\begin{abstract}
 We formalize networks with evolving structures as temporal networks and propose a generative link prediction model, Generative Link Sequence Modeling (GLSM), to predict future links for temporal networks. GLSM captures the temporal link formation patterns from the observed links with a sequence modeling framework and has the ability to generate the emerging links by inferring from the probability distribution on the potential future links. To avoid overfitting caused by treating each link as a unique token, we propose a self-tokenization mechanism to transform each raw link in the network to an abstract aggregation token automatically. The self-tokenization is seamlessly integrated into the sequence modeling framework, which allows the proposed GLSM model to have the generalization capability to discover link formation patterns beyond raw link sequences. We compare GLSM with the existing state-of-art methods on five real-world datasets. The experimental results demonstrate that GLSM obtains future positive links effectively in a generative fashion while achieving the best performance (2-10\% improvements on AUC) among other alternatives.
\end{abstract}

\section{Introduction}

Many real-world applications could be modeled as link prediction problems. For example, the recommendation system could be treated as a network system learns to connect user nodes with product nodes \cite{esslimani2011densifying}; the recommendation of friends in the social media is the prediction for the future links based on the current social network structure \cite{liben2007link}; even the financial risk could be discussed through the link formation probabilities between the financial organizations in an economic network \cite{10.1007/978-3-319-73198-8_10}. Two mainstream categories in link prediction are either based on the statistical patterns of the link formation behaviors of the network \cite{0295-5075-61-4-567,adamic2003friends,liben2007link} or the graph representation learning \cite{Tang:2015:LLI:2736277.2741093,DBLP:conf/aaai/WangWWZZZXG18} methods which embed nodes as vectors with respect to the network topological information. Most of these methods are discriminative models that verify whether an unknown link given during the test time is rational by training a classifier on existing links and negative samples \cite{menon2011link}. These methods show moderate performance by learning the decision boundary between positive samples (usually the observed links) and negative samples (usually random links between two arbitrary nodes). The temporal information of how links appear in a chronological order, which embodies rich information and useful in practical applications, is completely ignored. To further improve the modeling ability, recent works study the temporal link prediction \cite{Dunlavy:2011:TLP:1921632.1921636} which improve the prediction performance based on the temporal information captured by the time-dependent methods \cite{DBLP:conf/wsdm/YuA017}. However, since they do not consider the contextual relationship \cite{DBLP:journals/neco/WorgotterP05} contained in the chronological link sequence, they hardly capture the accurate network formation dynamics \cite{dynamics2005} (or link formation patterns) for the future links. The ignorance of the chronological link sequence during the formation progress of the evolving networks (e.g. the evolving social network \cite{Kossinets88} and evolving economic network \cite{Kirman1997}) brings the following two challenges for the temporal link prediction.
\begin{itemize}
 \item \textbf{Network dynamics.} Most classic methods based on the node-level empirical statistical rules \cite{0295-5075-61-4-567,adamic2003friends} without considering the network formation dynamics. This may result in the performance degradation when the statistical rules vary from time to time.
  \item \textbf{Network model bias.} Since it is difficult to model the link formulation patterns, the graph representation methods model the general latent link patterns from the observed network without considering the chronological order for links they are observed. Therefore, they hardly capture the link formation patterns directly and the network reconstructed by them with the historical data may bias from the current network, while the structure is already evolved with the new links added \cite{DBLP:conf/wsdm/YuA017}. This hampers the accuracy of the prediction results for the graph representation learning methods.
\end{itemize}

One way to solve the aforementioned challenges is to sort the links as a link sequence in their emerging time order and learn the link formation patterns based on the obtained link sequence. Inspired by the framework of the neural language modeling \cite{NIPS2014_5346} which studies the contextual relationship between observed words and the succeeding word in NLP fields, we adopt the sequence modeling techniques for temporal link prediction. By analogizing the idea of neural language modeling in NLP to temporal link prediction in graph mining, we formalize the link formation pattern as a conditional probability distribution and propose a neural network model (Generative Link Sequence Modeling, GLSM) to learn the temporal link formation patterns from the chronologically ordered link sequences with an RNN \cite{DBLP:conf/interspeech/MikolovKBCK10} based sequence modeling framework. Unlike previous discriminative counterparts, GLSM introduces a generative perspective which not only models the existence of different links but also the order that they are observed. It first learns the conditional probability distribution between the preceding and succeeding sequences enumerated from the observed link sequence and then predicts the future links with a generating process to sample the potential future links based on the learned distribution.

However, simply adopting raw links for sequence modeling may lead to several issues. Since a link is a tuple of nodes which contains the binary relationship between a source and a destination node, this relationship is discarded when we directly encode the links as the unary tokens like the text tokens in NLP models. Besides, too specific tokens, e.g. each one corresponding to a raw link between every two nodes, break down the dependencies among links with similar behaviors in the resulted token sequence. This may lead to the serious overfitting problem and thus the RNN could hardly capture any useful contextual relationship. To obtain the suitable token sequence for the sequence modeling framework, we propose a self-tokenization mechanism to control the granularity of the obtained tokens and the degree of contextual correlation in the resulted sequence automatically. The self-tokenization mechanism consists of a clustering process to obtain the abstract aggregation token alphabet and a mapping process to generate the tokens based on the resulted alphabet. With a differentiable clustering distance function, the loss function of the self-tokenization is incorporated into the loss function of the sequential modeling so that the model not only learns to self-tokenize the raw link sequences that preserve inter-link dependencies but also encode the temporal information among self-tokenized sequences via sequential modeling.

In the experiment sections, we verify that GLSM captures the useful contextual relationship between the preceding and succeeding link sequences and the generated future links cover the ground-truth positive links effectively in five real-world temporal networks. We also compare GLSM with the state-of-art methods on temporal link prediction tasks with different parameters, where GLSM outperforms existing methods. What's more, the experiment results in the case study also indicates that the self-tokenization mechanism helps GLSM capture the link formation patterns between the different communities of the temporal networks.

In summary, this work includes the following contributions:
\begin{itemize}
  \item We introduce temporal link prediction by a sequence modeling framework to discover the conditional distribution (defined as the temporal link formation pattern) between the preceding and the succeeding link sequences.
  \item We propose a self-tokenization mechanism to encode links as the tokens with respect to the clusters obtained by a clustering process on the original network while keeping the chronological order. This mechanism allows our method to capture the correct network formation dynamics from the observed network and thus alleviates the network model bias problem.
  \item We propose a two-step sampling link generator to generate potential future links based on the learned temporal link formation patterns from the observed network.
  \item We compare GLSM with state of the art methods on five real-world temporal network datasets and the results show that sequence modeling, along with the proposed self-tokenization mechanism, could achieve the best performance on the temporal link prediction tasks.
\end{itemize}

\section{Preliminary}

In this section, we formalize the related notations about the temporal network and define the temporal link prediction problem with the sequence modeling learning framework.

\subsection{Temporal Network}
We model the temporal network as a graph $G=\langle V, E(T)\rangle$ $(T>0)$ with a fixed node set $V$ and a link sequence $E(T)$. $T$ is the observing time for $G$, and the links in $E(T)$ are sorted in their emerging time order of the physical world and each link $e_t$ ($\forall e_t\in E(T)$) is a tuple $(u, v)$ where $u$ and $v$ are the nodes from the set $V$ and $t$ is the timestamp when $e_t$ is emerging between $u$ and $v$ ($t\in [0,T]$). Figure 1 illustrates an example of the temporal network in this link sequence form, where $V=\{1,2,3,4,5,6,7,8,9\}$ and $E(6)=\{e_1,e_2,e_3,e_4,e_5,e_6\}$, $E(8)=\{e_1,e_2,e_3,e_4,e_5,e_6,e_7,e_8\}$.
\begin{figure}[h]
\centering
\setlength{\fboxrule}{0.3pt}
  \setlength{\fboxsep}{0.1cm}
  \fbox{\includegraphics[width=2.4in]{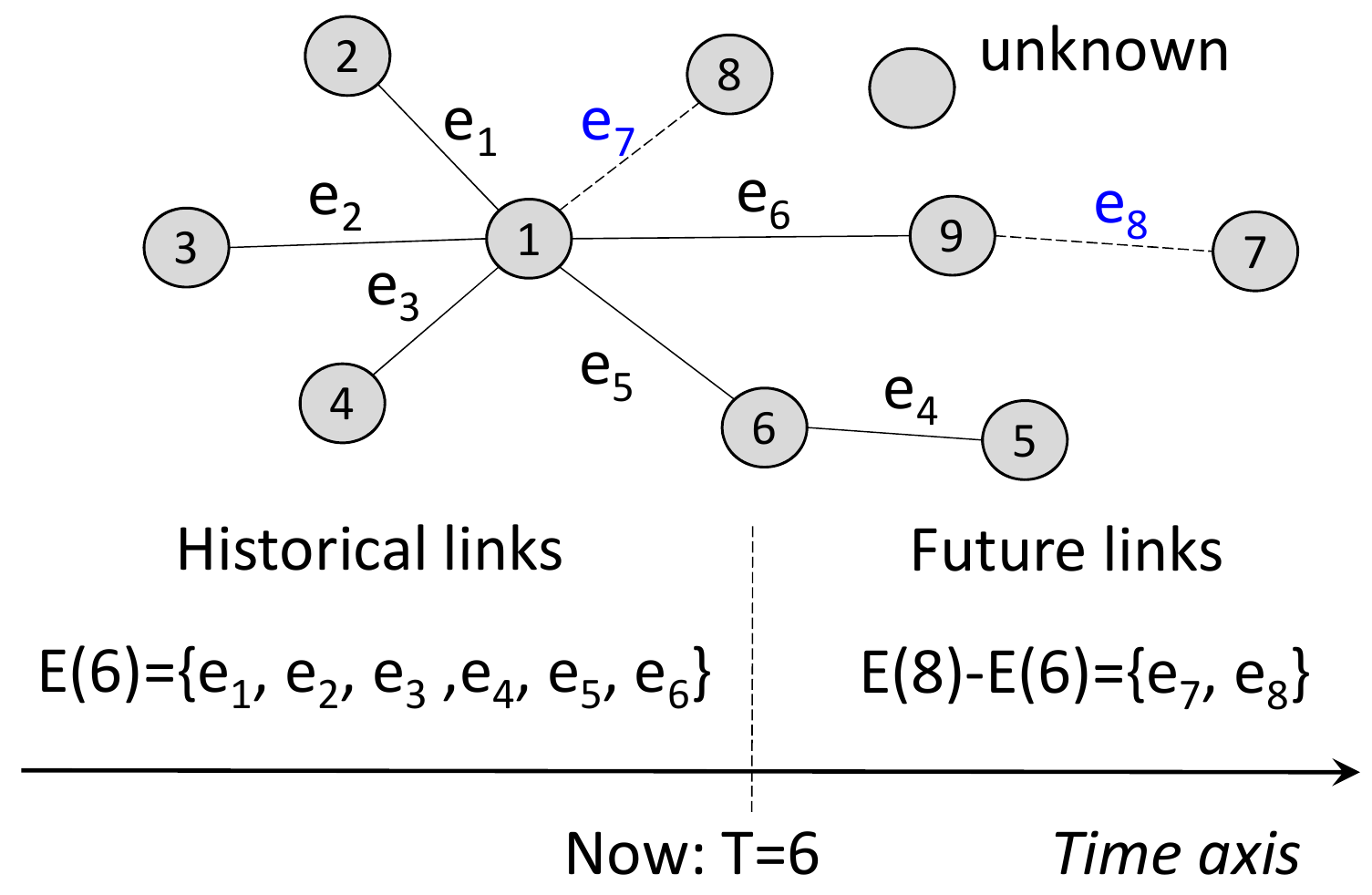}}
  \caption{Example of the temporal network}
   \label{fig:boxed_custom}
\end{figure}

In this setting, for not losing the generality, the newly emerging nodes could be treated as an ``unknown'' node which is also included in the node set $V$.

\subsection{Temporal Link Prediction}
In the practical application scenario, the intuitive requirement to predict the temporal links is to predict the potential future links given the historical links (e.g. given the current purchasing records in an e-commerce system, how to predict the future purchasing for users?). To solve the intuitive requirement in this scenario, we divide the temporal link prediction into two steps. First, to learn the temporal link formation pattern for the links from the historical link sequence, and then to infer the future links based on the learned patterns.

Suppose the observed links for the temporal network are the link sequence sorted in the chronological order, then the temporal link formation pattern describes the probability to observe a succeeding link sequence based on a preceding link sequence. We formalize the temporal link formation pattern in Definition 1.

\begin{definition}\rm
Temporal Link Formation Pattern. Given a temporal network $G(T)=\langle V,E(T)\rangle$ at time $T$, the temporal link formation pattern is defined as the conditional probability distribution $p(E(T)|E(0))$ which means the emerging probabilities for the links in $E(T)$ based on the initial link sequence $E(0)$. This conditional probability is computed in the following way.
\begin{equation}
p(E(T)|E(0))=\prod_{t=1}^{T} p(E(t)|E(t-1)).
\end{equation}
\end{definition}

With the definition for the temporal link formation pattern, the problem to learn the temporal link formation patterns from the observation is formalized as the following.

\begin{definition}\rm
Temporal Link Formation Pattern Learning. Given a temporal network $G(T)=\langle V,E(T)\rangle$ at time $T$, the temporal link formation pattern learning can be defined as the problem to estimate the conditional probability distribution $p(E(T)|E(0))$ through the optimization of the following Equation.
\begin{equation}
\arg\min\sum_{t=1}^{T}-p(E(t)|E(t-1))\log(p(E(t))),
\end{equation}
where $p(E(t)|E(t-1))$ is the estimated probability for the link sequence $E(t)$ based on the preceding sequence $E(t-1)$; $p(E(t))$ is the probability of $E(t)$ which is measured from the observation.
\end{definition}
The estimated probability $p(E(t)|E(t-1))$ also simplifies as $p(E'(t))$ in the remaining part.
Note that Equation (2) is a cross entropy function \cite{deBoer2005} to get the difference between the estimated probabilities and the observed probabilities. In this work, Equation (2) actually allows a model to learn the evolving pattern between the sequence of estimated future links and the sequence of historical links. Therefore, with the objective function in Equation (2), the learned temporal link formation pattern captures both rules of the network formation dynamics and the evolving structure for networks and thus alleviate the network model bias problem.
This is a sequence modeling problem \cite{NIPS2014_5346} while RNN is good at dealing with such a problem. Therefore, we propose an RNN based neural network model to solve it. After the training process of the RNN, we enumerate the future links based on the learned temporal link formation pattern.


\section{Our Framework}

In this section, we propose the Generative Link Sequence Modeling (GLSM) to learn the temporal link formation pattern and generate the future links. As shown in Figure 2, the GLSM consists of a training process to learn the temporal link formation patterns and a generating process to generate the future links. We will give the details of the GLSM in the remaining part.

\begin{figure}[h]
  \centering
  \subfigure[Training process of GLSM]{
    \label{fig:subfig:a} 
    \includegraphics[width=3.3in]{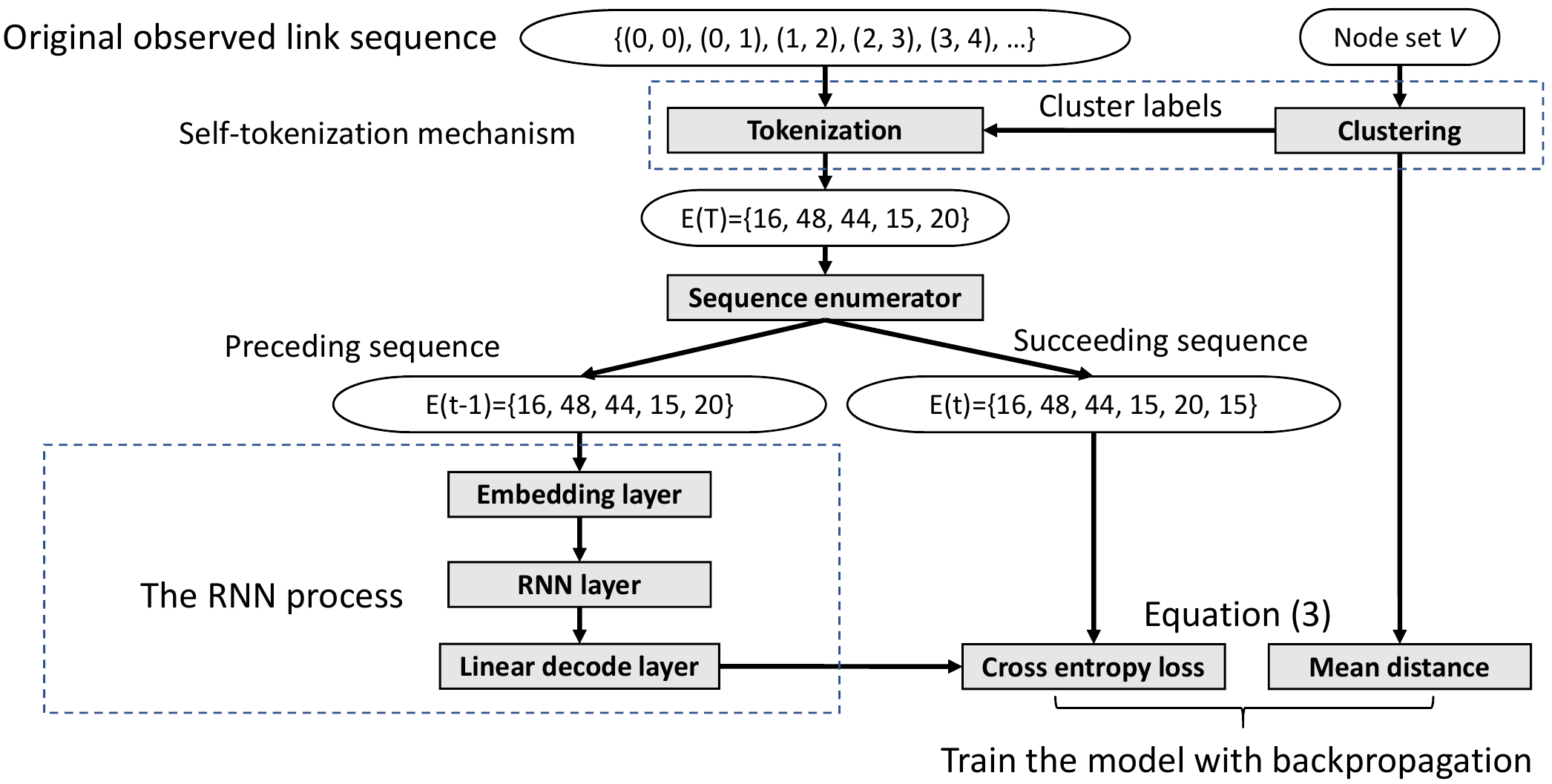}}
  \subfigure[Generating process of GLSM]{
    \label{fig:subfig:b} 
    \includegraphics[width=2.in]{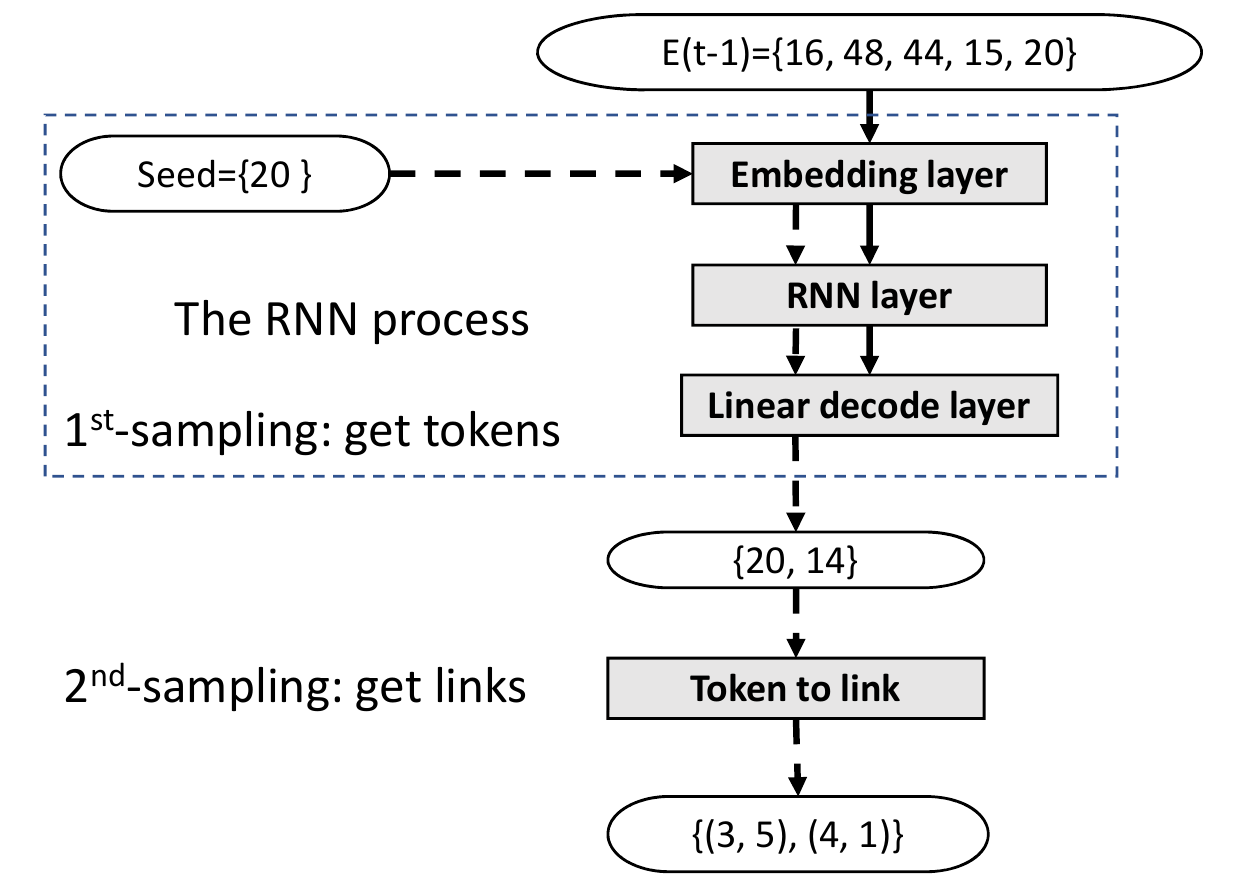}}
  \caption{Framework of GLSM}
\end{figure}
\subsection{Temporal Link Sequence Modeling via Self-tokenization}

We introduce the training process in Figure 2 (a) which learns the temporal link formation patterns with the sequence modeling framework in this section.

The key to implement the temporal link prediction with the sequence modeling framework is to convert the link sequences to a unary token sequence fitted the input of RNN. We formalize this process in following definitions.
\begin{definition}\rm
 Basic link sequence tokenization. Given a temporal network $G(T)$$=$$\langle$$V,E(T)$$\rangle$, the target of tokenization is to establish a tokenization map $g$:$E(T)$$\rightarrow$$A$ and result in a new sequence $E$$=$$\{g(e_t)\}$ ($\forall e_t$$\in$$E(T)$), where $A$ is an alphabet containing all the tokens $g(e_t)$ ($\forall e_t$$\in$$E(T)$).
\end{definition}

A naive tokenization method is to map every ``node-to-node'' link to a unique token. It results in a $|V|$$\times$$|V|$ alphabet $A$ which consists of all the possible links between any two nodes in $V$. This leads to a serious problem: since all the tokens in the tokenized sequence are different, when the RNN is fed with such a sequence, it is easily overfitting and outputs the patterns without any connection between the preceding and succeeding sequences. The experimental results in Figure 4 verifies this analysis. To solve this problem, we propose the self-tokenization mechanism which generates the token sequence with a clustering process.
\begin{definition}\rm
Self-tokenization Mechanism. Given a temporal network $G(T)$ $=$ $\langle$$V,E(T)$$\rangle$, suppose $C$ is the set of the obtained subgraphs (or communities) in $G(T)$ after the clustering of $G(T)$ ($\forall$$c_1$,$c_2$$\in$$C$, $c_1$$\cap$$c_2$$=$$\phi$). The target of self-tokenization is to automatically establish a tokenization map $g$:$C\times C$$\rightarrow$$A$ and result in a unary token sequence $E$$=$$\{g(e_t)\}$ ($\forall$$e_t$$\in$$E(T)$, $\forall g(e_t)\in A$), where $A$ is an alphabet containing all the tokens $g(e_t)$ ($\forall e_t$$\in$$E(T)$).
\end{definition}
 This tokenization method first applies a clustering to the network $G(T)$ to the different subgraphs (communities) and then maps the source and destination nodes of a specific link to the corresponding subgraphs (communities). This transforms the original ``node-to-node'' links to the ``community-to-community'' links. By constructing and referring to the alphabet $A$ with all of the resulted ``community-to-community'' links, we finally obtain a unary token sequence. For example, given a link sequence $E(3)=\{e_0,e_1,e_2\}$, where $e_0=(1,100)$, $e_1=(95,43)$ and $e_2=(78,25)$, suppose the clustering generates $3$ communities. The nodes $1$, $43$ and $78$ are in the $1$-st community, the node $95$ is in the $2$-nd community, and the node $25$ and $100$ are in the $3$-rd community, the original links can convert to $g(e_0)=(1,3)$, $g(e_1)=(2,1)$, and $g(e_2)=(1,3)$. Then the alphabet $A=\{1:(1,3),2:(2,1)\}$. Finally, the original link sequence is tokenized to $E=\{1,2,1\}$ according to $A$.

This mechanism applies the hypergraph partitioning method \cite{1597385} which is effective in controlling the granularity of the graph for the graph mining methods. Therefore, the resulted token sequence is coarser in granularity and more general than the original link sequence. Furthermore, since the clustering results in the non-overlapping communities which consist of similar nodes, the resulted sequence also contains the network topological information. After the clustering process, the ``unknown'' nodes can also be labeled as a single community before the sampling process.

With this setting, the number of the community $|C|$ is used as a parameter to control the size of the output alphabet $A'$ and thus determine the generality of the resulted token sequence. With the special conditions, with $|C|=|V|$, the self-tokenization mechanism can be reduced to the basic link sequence tokenization in Definition 3.

To generate the suitable communities (or subgraphs) for this task, we integrate a self-clustering process \cite{DBLP:journals/corr/abs-1802-03997} in the Equation (2) and result in the following loss function.
\begin{equation}
\begin{split}
\arg\min(1-\alpha)\sum_{t=1}^{T'}-p(E(t)|E(t-1))\log(p(E(t)))\\+\alpha\sum_{c\in C}\sum_{v\in c}d(v,u(c)),
\end{split}
\end{equation}
where $C$ is the set of communities, $u(c)$ is the center node of the $c$-th community, $d(*)$ can be any differentiable clustering distance function between two nodes and $\alpha$ is the weight of the clustering loss. Note that the links in Equation (3) have already been transformed as the ``community-to-community'' tokens which are different from the ``node-to-node'' links in Equation (2). As it is illustrated in Figure 2 (a), where tokenization for the sequences is based on the obtained communities from the clustering process, the temporal link formation patterns and the clustering are learned simultaneously.

The pseudocode of the forward algorithm about the temporal link formation modeling, which  corresponds to the training part indicated in Figure 2 (a), is listed in Algorithm 1.
\begin{algorithm}[h]\small
\DontPrintSemicolon \KwData{Temporal network $G(T)=\langle V,E(T)\rangle$, cluster number $K$, clustering weight $\alpha$, epoch threshold $N$, chunk size $H$, stride $s$}
\KwResult{The trained RNN, the tokenization map $g$}
\Begin {
Initialize the hidden state $h(0)$\\
\For{epoch$\in [0, N]$}{
Compute the map $g$ and the average clustering distance $\bar{d}$ for all $v$ in $V$ with $K$ center nodes\\
Generate the token sequence $E$ given $E(T)$ and $g$ (Def. 3.2)\\
Enumerate a chunk $E(T^*)$ from $E$ with size $H$ randomly\\
Get the preceding sequence $E(t-1)$ and the succeeding sequence $E(t)$ from $E(T^*)$ given $s$\\
Infer the conditional distribution $p(E'(t))$ by Eq. (4)\\
Compute the loss $l$ given $p(E'(t))$, $p(E(t))$ by Eq. (2)\\
Compute the final loss $L$ given $l$, $\bar{d}$ and $\alpha$ by Eq. (3)\\
}
\textbf{Output} the final loss $L$, the tokenization map $g$
}
\caption{Temporal link sequence modeling}
\end{algorithm}
Line 4 clusters the nodes in $K$ different clusters and Line 5 tokenizes the original link sequence to the unary token sequence with the self-tokenization mechanism. To learn the temporal link formation patterns from the tokenized sequences, for each epoch, Lines 6 and 7 generate a preceding sequence $E(t-1)$ and a succeeding sequence $E(t)$ randomly from the input sequence, where the stride $s$ is a parameter to control the overlapped portion between $E(t-1)$ and $E(t)$. After that, Line 8 computes the conditional distribution $E'(t)$ with the RNN by Equation (4).
\begin{equation}\small
p(E'(t)),h(t)\leftarrow RNN(p(E(t-1)),h(t-1)),
\end{equation}
where $h(t)$ is the hidden state value at time $t$.
Line 9 computes the cross entropy loss with $p(E'(t))$ and $p(E(t))$ by Equation (2). The final loss $L$ is calculated by Equation (3). Since Equation (3) incorporates the cross entropy and the clustering loss (distance) together, this model trains the RNN and performs the clustering simultaneously. The training applies the backpropagation with the Adam optimizer \cite{DBLP:journals/corr/KingmaB14}. What's more, since the complexity of Algorithm 1 is proportional to the epoch threshold, its time complexity is $O(N)$.

When the training process is complete, We obtain a trained RNN containing the temporal link formation patterns and the tokenization map $g$ which records the labels for all the nodes in $V$ respectively. Since the obtained RNN captures the network formation dynamics, we can generate the future possible network structure as the prediction for the evolving network with it and this alleviates the network model bias problem.

\subsection{Generating Links with the Two-step Sampling Link Generator}
\begin{algorithm}[h]\small
\DontPrintSemicolon \KwData{a trained RNN and tokenization map $g$, the link sequence $E(T)$, generation round $R$}
\KwResult{a set of predicted positive links $\Delta E$}
\Begin {
Generate the token sequence $E$ given $E(T)$ and $g$ (Def. 3.2)\\
Randomly enumerate a 1-length sequence $E(0)$ from $E$\\
Initialize the hidden states $h(0)$\\
\For{$t\in[1,R]$}{
Infer the conditional distribution $p(E'(t))$ with $E(t-1)$ and $h(t-1)$ by Eq. (4)\\
$1^{st}$-sampling: draw the next token $e$ from a multinomial distribution experiment given $p(E'(t))$\\
$2^{nd}$-sampling: draw the next link $(u,v)$ given $e$\\
\If{$(u,v)\notin E(T)\bigcup\Delta E$}{
$\Delta E\leftarrow\Delta E\bigcup(u,v)$}
$E(t)\leftarrow E(t-1)\bigcup\Delta E$
}
\textbf{Output} $\Delta E$
}
\caption{Two-step sampling link generator}
\end{algorithm}
With the trained RNN, we propose the two-step sampling link generator to generate the new links and this process is shown in Algorithm 2. This corresponds to the generating process in Figure 2 (b). Its basic idea is to sample the links according to the conditional probabilities learned by the trained RNN. Since the token obtained by the self-tokenization mechanism refers to the abstract ``community-to-community'' link, Algorithm 2 consists of two sampling steps to get the ``node-to-node'' links. The first sampling step is started with a randomly chosen ``seed'' link sequence in Line 3. With the seed link sequence, Line 6 iteratively infers the next probability distribution $p(E'(t))$ for all the tokens of the alphabet $A$. Line 7 generates the next token $e$ according to $p(E'(t))$ with a multinomial distribution sampling without replacement.

\textbf{$2^{nd}$-sampling.} Since $e$ is actually the token of a specific ``community-to-community'' link, to obtain a ``node-to-node'' link which may appear in the original network, Line 8 samples the ``node-to-node'' link with a given token $e$. The sampling is implemented through drawing source and destination nodes from the two related communities of $e$ respectively. Note that this $2^{nd}$-sampling is a general process which could use any sampling method such as the weighted random sampling, the greedy sampling, the beam sampling, etc. The candidate link set is generated by enumerating a complete combination of all the possible links between the nodes in the source community and the destination community for a related token $e$. The linkage probabilities are computed by Equation (5).
\begin{equation}\small
p(V_1,V_2)=f(V_1)\cdot{f(V_2)}^T,
\end{equation}
where $V_1$ and $V_2$ are the node sets of the source and destination communities related to $e$ and $f(*)$ is the embedding layer of the nodes trained by the clustering process in Algorithm 1. $f(*)$ contains the latent features for all the nodes, and a multiplication of two embedding vectors in Equation (5) actually computes all the linkage probabilities for the related nodes between the two corresponding communities. This process prunes the search space for the link probabilities of the next links and incorporates the temporal information into the basic graph representing framework. We verifies that this method indeed improves the quality of the generated links in the experiment. For each iteration of Algorithm 2, a newly generated link which is not included in $\Delta E$ and the existing link set $E(T)$ will be appended at the end of the link sequence $\Delta E$ and it is also used as the input for the next sampling. The complexity of Algorithm 2 is $O(R)$ which is positively proportional to the round number $R$.

\section{Experiment and Discussion}

\subsection{Dataset}

We compare our methods with the existing methods on five real-world datasets. Their details are shown in Table 1.
\begin{table}[h]\small
\centering
\begin{tabular}{lccccc}
\toprule
&CollegeMsg&Movielens&Bitcoin&AskUbuntu&Epinions\\\midrule
Total links&59,835&100,000&35,592&100,000&100,000\\
Source card.&1,350&944&4,814&10,016&6,718\\
Destiny card.&1,862&1,683&5,858&10,001&22,406\\
Node number&1,899&1,682&5,881&12,513&27,370\\
Edge density&0.024&0.126&0.0025&0.0020& 0.0004\\
Rating range&0-1&0-5&0-20&0-1&0-5\\
Days covered&193&214&1,903&1,038&4,326\\
Start&2004.04&1997.09&2010.11&2001.09&1999.07\\
End&2004.10&1998.04&2016.01&2003.06&2011.05\\
\bottomrule
\end{tabular}
\caption{Dataset statistics}
\end{table}

Our datasets cover the different applications on the recommendation system and social network. ``Movielens''\footnote{https://grouplens.org/datasets/movielens/}, ``Netflix''\footnote{https://www.kaggle.com/netflix\-inc/netflix\-prize\-data} and ``Epinions'' \cite{tang-etal12b} are the classic datasets to test the performance of link prediction or recommendation models. ``CollegeMsg'' is a binary online social network from \cite{DBLP:journals/jasis/PanzarasaOC09}. ``Bitcoin'' is from \cite{kumar2016edge} and it records the trust scores between the users from the Bitcoin online marketplaces\footnote{https://bitcoin\-otc.com/} in the corresponding transactions. All dataset are in the format of ``source, destination, rating, timestamp''. All ratings are adjusted to 0 or 1 since our method can only deal with the binary prediction problem. The node number is the number of different nodes after merging the same nodes in source and destination position of the records.

\textbf{Preprocessing for Temporal Link Prediction Task}

To test the performance of temporal link prediction for all the methods, we preprocess the datasets in the following way.
\begin{itemize}
\item To simulate the link formation in the real scenario, we first order the links in each dataset in chronological order separately.
\item Then, for each dataset, we divide the ordered links into $N$ chunks with equal time-span and conduct the training-and-testing within each chunk. The links are also in chronological order in each chunk. The statistical results across all chunks of a dataset produce the mean and stand-deviation results for each method.
\item We select a ratio, training ratio $\gamma$, of links as the training set (historical links) and leave the remaining links as the testing set (future links). With this setting, we test the performance of state of the art methods in predicting the real ground-truth future links.
\end{itemize}

\subsection{Experiment settings and benchmark.}

\textbf{Comparison methods.}

Our methods are compared with the state-of-the-art link prediction methods which are used in most related studies.
\begin{itemize}
  \item Jaccard Coefficient (JC) \cite{liben2007link} and Adamic Adar (AA) \cite{adamic2003friends}. JC and AA are the classic link prediction methods based on the statistical similarity scores \cite{BARABASI2002590}. They assume that, for a social network, two unconnected nodes with the high statistical scores have the large probability to be linked together in the future. Their similarities are computed based on the common neighbor number \cite{Newman404} between two nodes in different forms.
  \item Matrix Factorization (MF) \cite{10.1007/978-3-642-23783-6_28}. MF factorizes the adjacency matrix of the network into two matrices with latent feature dimension. Since it's factorized matrices could easily be explained with the relationship between the latent user features, it is applied in many real-world network systems as the recommendation algorithm.
  \item Temporal Matrix Factorization (TMF) \cite{DBLP:conf/wsdm/YuA017}. TMF uses the time-dependent matrix factorization method to improve the temporal link prediction performance from the original MF method.
  \item Graph GAN (GG) \cite{DBLP:conf/aaai/WangWWZZZXG18}. GG is a neural network model which is based on the framework of the graph representing method and GAN \cite{NIPS2014_5423}. With the dynamic game between the generator and discriminator, this model could reach higher performance than the previous methods such as LINE \cite{Tang:2015:LLI:2736277.2741093}, DeepWalk \cite{Perozzi:2014:DOL:2623330.2623732}, etc. Therefore, we pick GG as the representative method for the graph representation learning methods.
  \item Graph AutoEncoder (GAE) \cite{DBLP:journals/corr/KipfW16a}. GAE applies the Graph Convolution Neural Network (GCN) to capture the topological information of graphs and then represents the basic graph as feature vectors. When doing the link prediction task, GAE reconstructs the graph with the obtained vectors and checks the link possibilities between corresponding nodes. Since the GCN-captured feature vectors collect more topological information than basic graph embedding methods, GAE performs better than the basic methods.
  \item Link Prediction based on Graph Neural Network (LPGNN) \cite{DBLP:conf/nips/ZhangC18}. LPGNN first applies embedding methods to encode original graphs and then feeds the obtained latent features into a GCN layer. This method also outperforms the GAE.
\end{itemize}

\textbf{Experiment settings.}

To make the comparison fair, we implement all these methods in our prototype system with the GPU-version Pytorch and thus these methods are compared with the same data and prediction task platform. In our system, each method generates a link set with the corresponding emerging probabilities for the links. For our method, GLSM, after it generates future positive links, we also sample the negative future links for GLSM from the subtraction set of the current non-existing links and the generated positive links. The set of the negative links is set to the same size of the positive links.

During all the experiments, GLSM uses a Long Short-Term Memory neural network (LSTM) \cite{DBLP:conf/interspeech/SundermeyerSN12} version of RNN with 128 hidden states and 2 layers. We set the clustering weight $\alpha$ to 0.5 and use the k-means clustering for the self-tokenization process. For the $2^{nd}$-sampling process of GLSM, we use the weighted random sampling since it is the easiest to implement. All results of GLSM are generated after the training process with 20 epochs.

Our experiment is twofold. First, we compare all mentioned methods on five real-world datasets with temporal link prediction tasks in different parameters. Then, we perform two case studies to analyze the practical meaning of the obtained clusters after training.

\subsection{Effectiveness Experiment}

In this experiment, we split each dataset into different windows, and test the methods for all the windows. We first compare the prediction performances of the mentioned methods on the different datasets, and then, we analyze the hit ratios (measures the output quality) of the generated links for all methods on all the datasets. Finally, we analyze the sensitivity of GLSM in different parameters.

To further compare the prediction performance, we use the different training ratio $\gamma$ to test the capabilities of the methods to get the correct future links. Since the training ratio $\gamma$ decides the ratio of the data to be used as the training set (historical links) and the remaining data as the testing set (future links), the smaller the training ratio $\gamma$ is, the bigger the test set with the future links is. This shows that how far into the future could these methods predict with good performance.

\textbf{Temporal link prediction performance}

We compare the temporal link prediction performances of all the mentioned methods with ROC-AUC and RMSE (since the result of GLSM is binary, we set the weight to any link to 1 in RMSE test) on all the datasets with 10,000 link window size. ROC-AUC result shows the capabilities of the methods to distinguish positive and negative links and RMSE result shows the accuracy of the output linkage probabilities. Tables 2 and 3 illustrate the ROC-AUC and RMSE results. We set the training ratio $\gamma=0.7$ for all the datasets.

\begin{table}[h]\small
\centering
\scalebox{0.8}[0.8]{
\begin{tabular}{lccccc}
\toprule
Model&CollegeMsg&Movielens&Bitcoin&AskUbuntu&Epinions\\\midrule
GLSM&\textbf{0.7132$\pm$0.0031}&\textbf{0.7521$\pm$0.0007}&\textbf{0.7509$\pm$0.0016}&\textbf{0.7508$\pm$0.0001}&\textbf{0.7504$\pm$0.0012}\\
LPGNN&0.6773$\pm$0.0496&0.7109$\pm$0.0054&0.5796$\pm$0.0040&0.6762$\pm$0.0429&0.7109$\pm$0.0054\\
GAE&0.6634$\pm$0.0031&0.6931$\pm$0.0016&0.7372$\pm$0.0016&0.7241$\pm$0.0014&0.6591$\pm$0.0007\\
AA&0.5501$\pm$0.0003&0.7447$\pm$0.0006&0.6033$\pm$0.0002&0.6129$\pm$0.0023&0.5111$\pm$0.0000\\
MF&0.5395$\pm$0.0003&0.6055$\pm$0.0021&0.5854$\pm$0.0011&0.5717$\pm$0.0009&0.5123$\pm$0.0001\\
GG&0.5659$\pm$0.0006&0.5354$\pm$0.0006&0.5611$\pm$0.0013&0.5550$\pm$0.0018&0.5245$\pm$0.0003\\
TMF&0.5175$\pm$0.0001&0.5086$\pm$0.0001&0.5451$\pm$0.0002&0.5249$\pm$0.0002&0.5163$\pm$0.0002\\
JC&0.5390$\pm$0.0002&0.6748$\pm$0.0019&0.5791$\pm$0.0000&0.5960$\pm$0.0015&0.5110$\pm$0.0000\\
\bottomrule
\end{tabular}
}
\caption{AUC results with 10,000 link window}\vspace{-0.1in}
\end{table}\vspace{-0.1in}
\begin{table}[h]\small\vspace{-0.1in}
\centering
\begin{tabular}{lccccc}
\toprule
Model&CollegeMsg&Movielens&Bitcoin&AskUbuntu&Epinions\\\midrule
GLSM&\textbf{0.20$\pm$0.10}&\textbf{1.96$\pm$0.72}&\textbf{9.37$\pm$0.03}&\textbf{0.03$\pm$0.01}&\textbf{2.53$\pm$0.43}\\
GAE&0.25$\pm$0.00&2.60$\pm$0.04&9.97$\pm$3.46&0.31$\pm$0.01&2.56$\pm$0.20\\
AA&0.90$\pm$0.00&3.27$\pm$0.02&11.54$\pm$0.16&0.97$\pm$0.01&4.19$\pm$0.01\\
MF&0.58$\pm$0.00&3.02$\pm$0.03&11.09$\pm$0.24&0.56$\pm$0.01&3.71$\pm$0.01\\
GG&0.61$\pm$0.00&3.34$\pm$0.03&11.23$\pm$0.22&0.59$\pm$0.01& 3.71$\pm$0.01\\
TMF&0.66$\pm$0.00&3.23$\pm$0.04&11.22$\pm$0.23&0.67$\pm$0.02&3.74$\pm$0.01\\
JC&0.98$\pm$0.00&3.57$\pm$0.05&11.76$\pm$0.19&0.99$\pm$0.00&4.20$\pm$0.01\\
\bottomrule
\end{tabular}
\caption{RMSE results with 10,000 link window}\vspace{-0.1in}
\end{table}\vspace{-0.1in}

We observe from Tables 2 and 3, the GLSM achieves the best performance of all the methods and its results are stable even on the sparse datasets with the edge degree lower than 0.002.  This indicates that GLSM captures the temporal link formation pattern correctly, and the obtained patterns are indeed helpful in generating the potential future links.

\textbf{Performance comparison in different parameters.}

Figure 3 compares the ROC-AUC results of all the methods on the ``Epinions'' data with different window sizes and the training ratios respectively. The links in our datasets are ordered chronologically to simulate the temporal link prediction task. Since most of the baselines can not directly capture this information, they perform barely (with AUC around 0.5) in this experiment.
\begin{figure}[h]
  \centering
  \subfigure[Diff. window sizes]{
    \label{fig:subfig:a} 
    \includegraphics[width=2.1in]{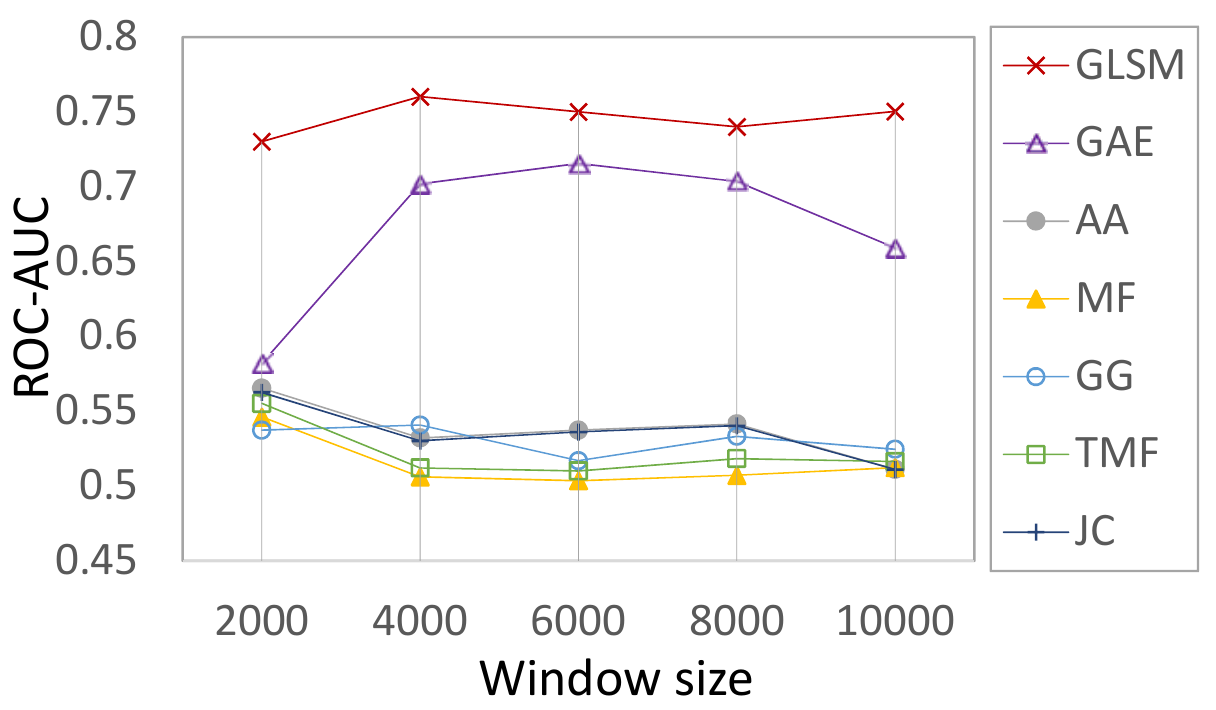}}
  \subfigure[Diff. training ratios]{
    \label{fig:subfig:b} 
    \includegraphics[width=2in]{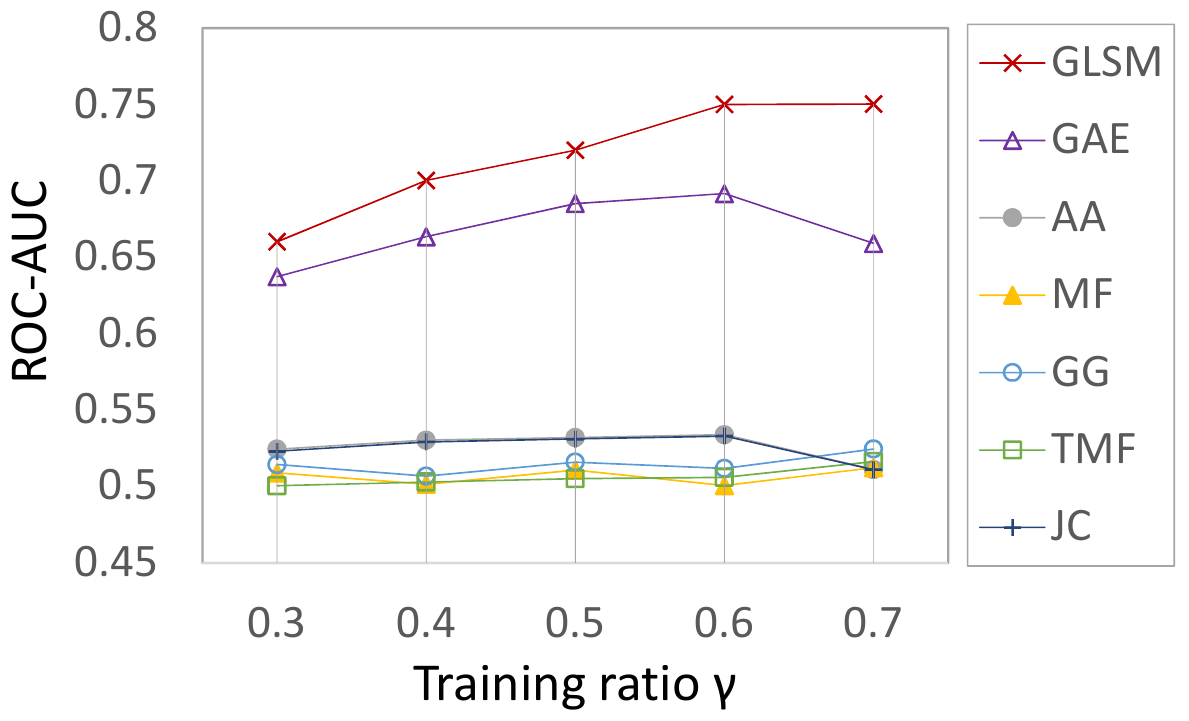}}
  \caption{Comparison in different parameters}
\end{figure}

We observe from Figure 3 that GLSM performs the best on all the different parameters. Figure 3 (a) indicates the sensitivities of the models to the different window sizes, and from Figure 3 (a), GLSM could reach a relative high AUC score even in the relatively small window size (2,000 links). Moreover, Figure 3 (b) demonstrates the abilities of the models to capture the future link patterns based on a small ratio of training data. In figure 3 (b), GLSM predicts the future links well with small training data ratio (under 0.5).

\textbf{Hitting ratio analysis.}

Since GLSM iteratively generates the future links which are different from the prediction scoring of the other methods, we analyze the quality of GLSM's generated links in different iteration rounds. We provide a metric, the hitting ratio, to measure the ratio of the captured real future links by the output of GLSM. This experiment runs on all the datasets with 10,000 links window size. The training ratio $\gamma$ is set to 0.7. The results are listed in Table 4, where $\overline{p_e}$ is the edge density of the network constructed by the testing data. In this setting, the hitting ratio for the random method equals to the corresponding $\overline{p_e}$ of a given dataset.
\begin{table}[h]\small
\centering
\begin{tabular}{cccccc}
\toprule
Iteration&CollegeMsg&Movielens&Bitcoin&AskUbuntu&Epinions\\\midrule
500&0.0054&0.0407&\textbf{0.0061}&\textbf{0.0054}&0.0007\\
1000&0.0063&0.0456&0.0049&0.0050&\textbf{0.0019}\\
1500&\textbf{0.0075}&\textbf{0.0499}&0.0042&0.0043&0.0016\\
2000&0.0053&0.0460&0.0035&0.0042&0.0017\\
2500&0.0044&0.0446&0.0025&0.0038&0.0009\\
3000&0.0043&0.0474&0.0030&0.0049&0.0006\\
\hline
$\overline{p_e}$&0.0041&0.0046&0.0015&0.0026&0.0002\\
\bottomrule
\end{tabular}
\caption{Comparison on hitting ratios in all datasets}
\end{table}

From Table 4, we observe that the hitting ratios of the generated links by GLSM significantly exceed the corresponding $\overline{p_e}$ and the hitting ratios reach the max within 1,500 iterations on all the datasets. This shows that the links generated by GLSM cover the true positive future links well.

\textbf{Sensitivity in different parameters.}

We analyze the influence of different cluster and chunk numbers on the AUC performance. This experiment is completed on the ``Movielens'' dataset within the 1,000 link window which involves 586 nodes. We set the epoch to 20 and iteration round to 1,500. The result is shown in Figure 4. In this experiment, GLSM reaches the best performance when the chunk number is between 50 and 100 and its performance degrades when the cluster number is under 50 or above 250. This addresses the discussion in 3.1 about the difference between the basic tokenization and the self-tokenization mechanism. That is, when the cluster number approximates the cluster number, the self-tokenization mechanism can be reduced to the basic method in Definition 3. What's more, the small cluster number leads to an alphabet with low generality which also degrades the prediction performance.

\begin{figure}[htbp]
\raggedright
\begin{minipage}[t]{0.5\textwidth}
\centering
\includegraphics[width=2in]{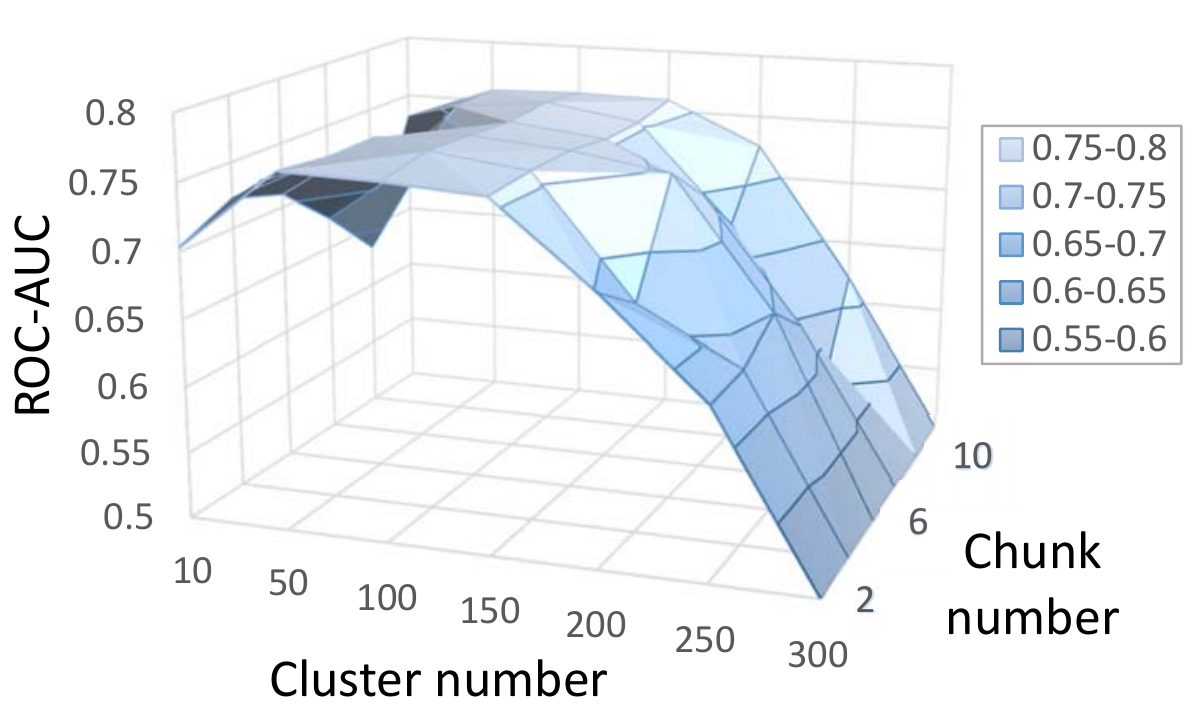}
\caption{Sensitivity}
\end{minipage}
\begin{minipage}[t]{0.49\textwidth}
\centering
\includegraphics[width=2in]{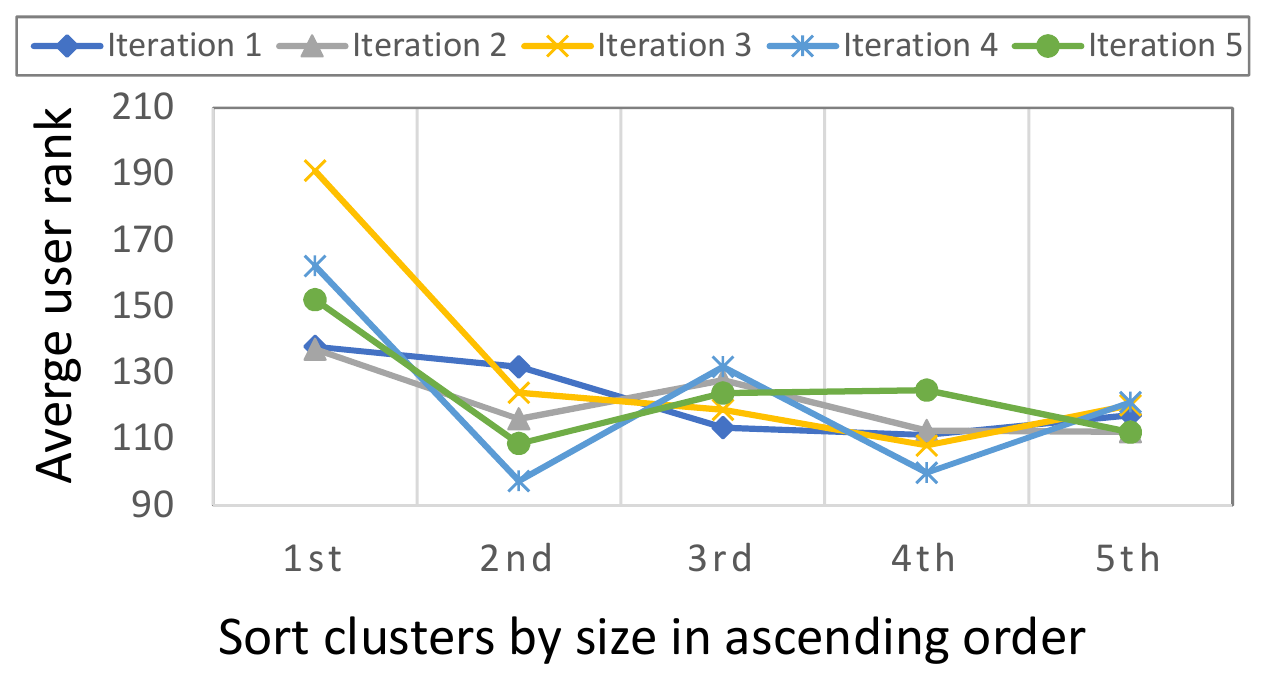}
\caption{Case study}
\end{minipage}
\end{figure}\vspace{-0.1in}

\subsection{Case study: Early User More Popular}

We analyze the clustering result obtained by the self-tokenization process of GLSM on the one-mode network from the ``CollegeMsg'' dataset. In this study, we order all users in ``CollegeMsg'' in the ascending order according to their registered time. Thus the early user has the high rank and vice versa. Then, we compute the average user rank for each cluster after the training of GLSM. We use the first 1,000 links of ``CollegeMsg'' network which relates to 237 users and set the clustering number to 5. Based on the settings, we analyze the relationship between the size of the cluster and the average user rank in the corresponding cluster. We show the results after five independent trainings in Figure 5. We observe that the average user ranks of the clusters are significantly different and the average user rank has a strong trend to diminish with the increase of the cluster size. Since the cluster size reflects the popularity for the corresponding cluster (community), this result agrees with the conclusion in \cite{doi:10.1002/asi.21015} that the early users (with small average user rank) in the social network are more popular. We further find that the variance of the average user rank in each cluster is smaller than the variance of the average user rank for all the users. This indicates that the registered time of the users in each cluster is close to each other. Therefore, the clustering result of the self-tokenization process helps GLSM capture the link formation patterns between the different communities of users (or nodes) for the temporal network.

\subsection{Case study: General Recommendation}

We analyze the clustering result obtained by the self-tokenization process of GLSM on the network from the ``Movielens'' dataset. In this study, we use the first 500 links of the ``Movielens'' dataset which relates to 286 nodes (including both the users and movies) and set the clustering number to 4. Since the ``Movielens'' is a bipartite network and our model can not directly be applied to the bipartite problem, we transform the bipartite network as a one-mode graph by labeling the nodes in different sets with unique identities. To keep the bipartite information, we retain the map between the obtained identities and the original nodes. Based on this map, we further divide the clustering results to the user communities and the movie communities. Therefore, we obtain 8 communities after the training process of GLSM. We highlight 2 user communities and 2 movie communities from the results in Figure 6 respectively. We observe that the users or movies with similar network topological structures are classified to the same communities. After the tokenization with this clustering results, in the generating process, the 1-st sampling process of Algorithm 2 will sample a token which is related to any two communities with different types (e.g. community 1 and community 2) and the 2-nd sampling process of Algorithm 2 will generate a specific ``user-to-movie'' link (recommendation) based on the selected communities from the 1-st sampling. This shows that the self-tokenization prunes the complete combination for all ``user-to-movie'' links by utilizing the network topological structure to get the candidate link set and then generate the specific links based on the obtained candidate link set. Consequently, besides generating the abstract token sequence, the self-tokenization mechanism also helps to improve the prediction performance for GLSM with the network topological information.

\begin{figure}[h]
  \centering
  \subfigure[Community 1]{
    \label{fig:subfig:a} 
    \includegraphics[width=2in]{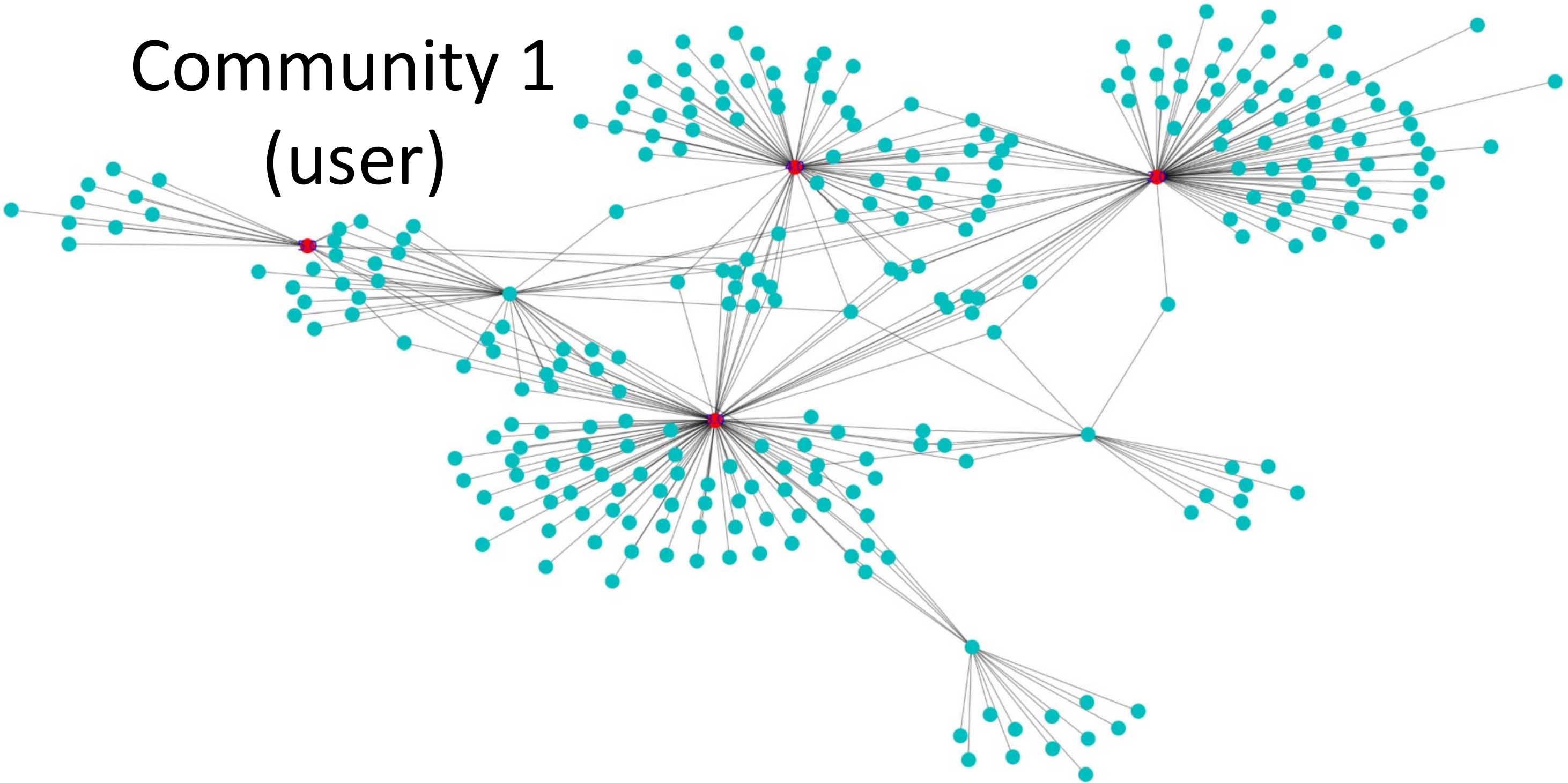}}
  \subfigure[Community 2]{
    \label{fig:subfig:b} 
    \includegraphics[width=2in]{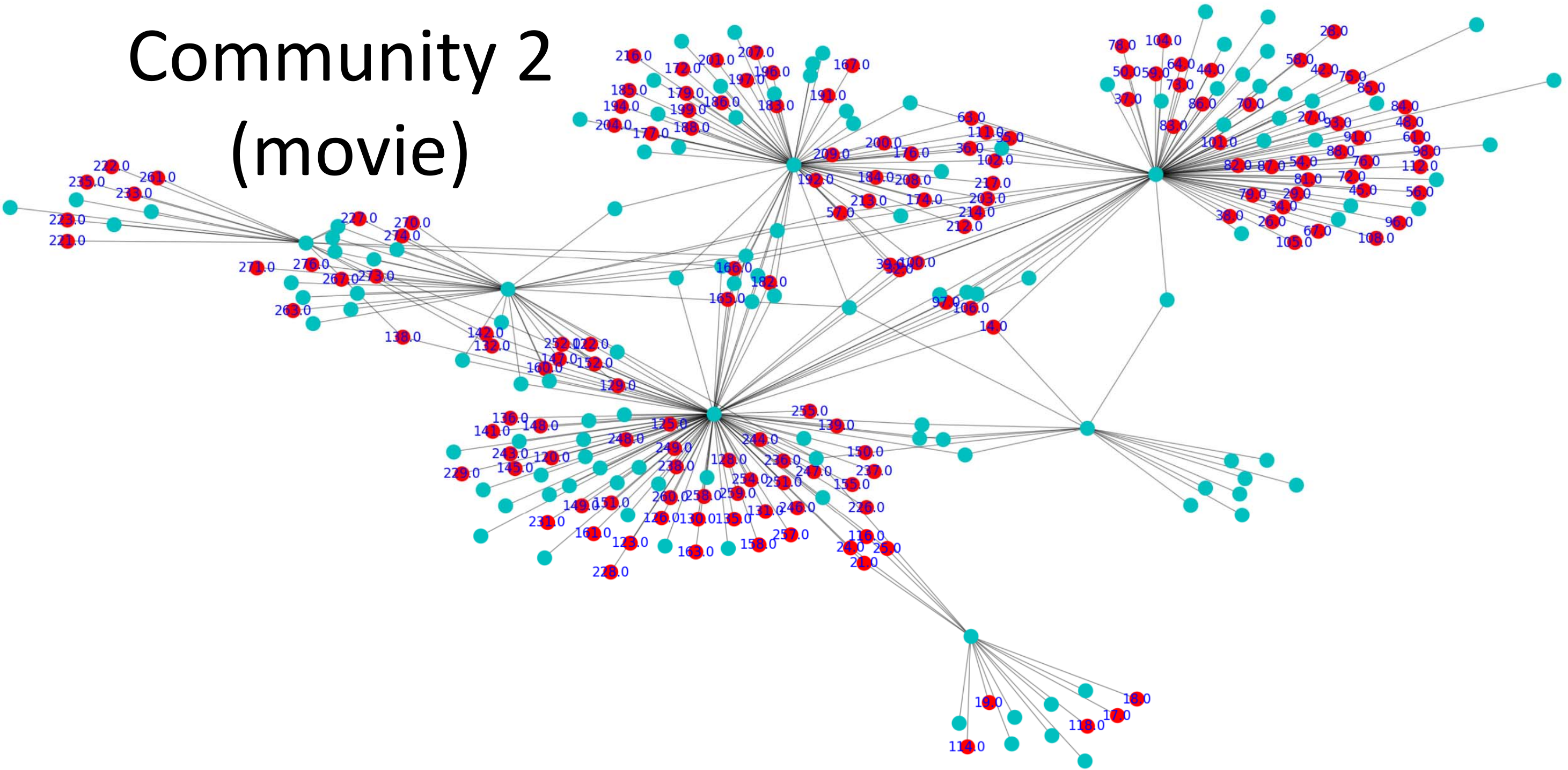}}
  \subfigure[Community 3]{
    \label{fig:subfig:b} 
    \includegraphics[width=2in]{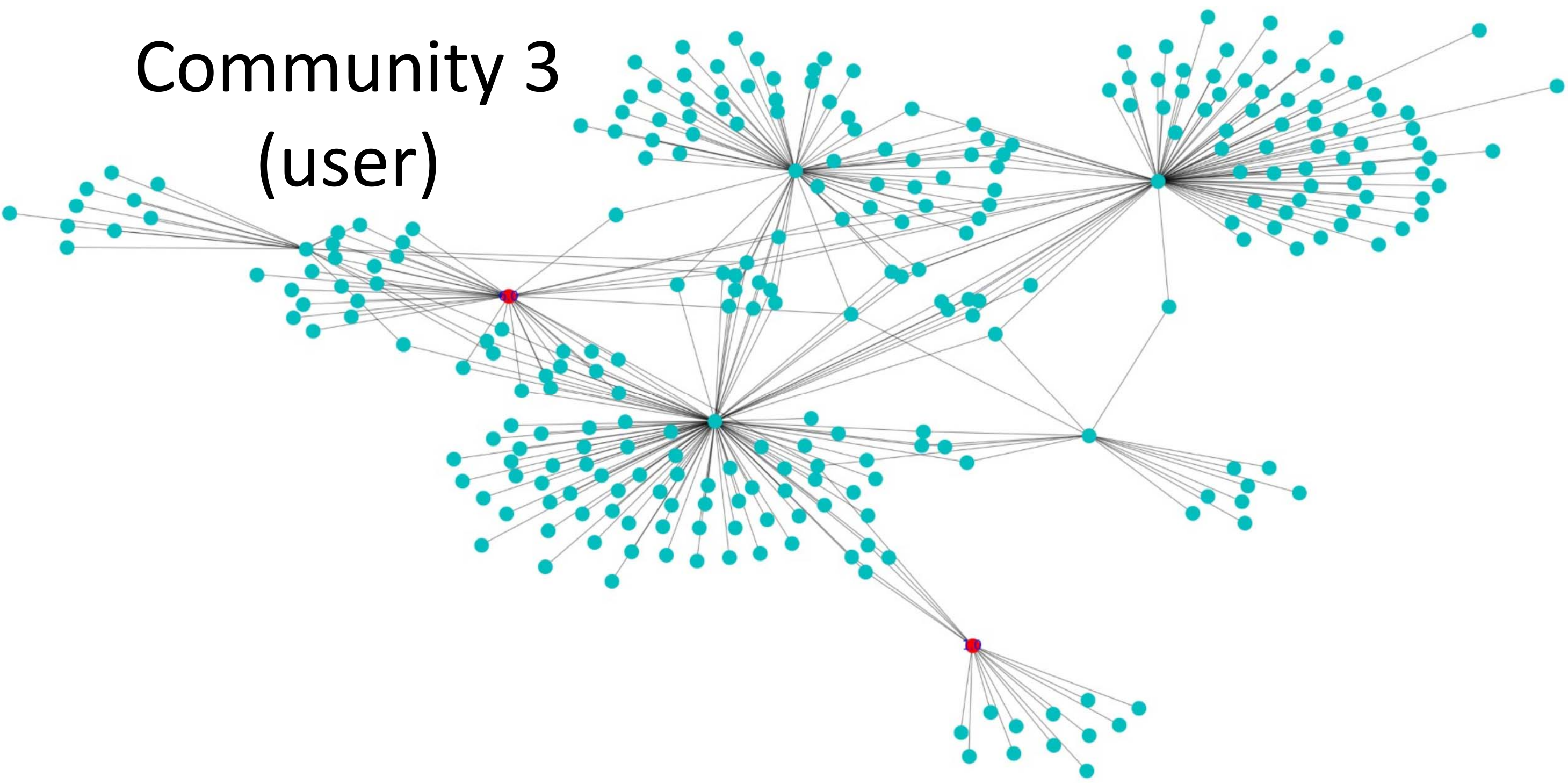}}
  \subfigure[Community 4]{
    \label{fig:subfig:b} 
    \includegraphics[width=2in]{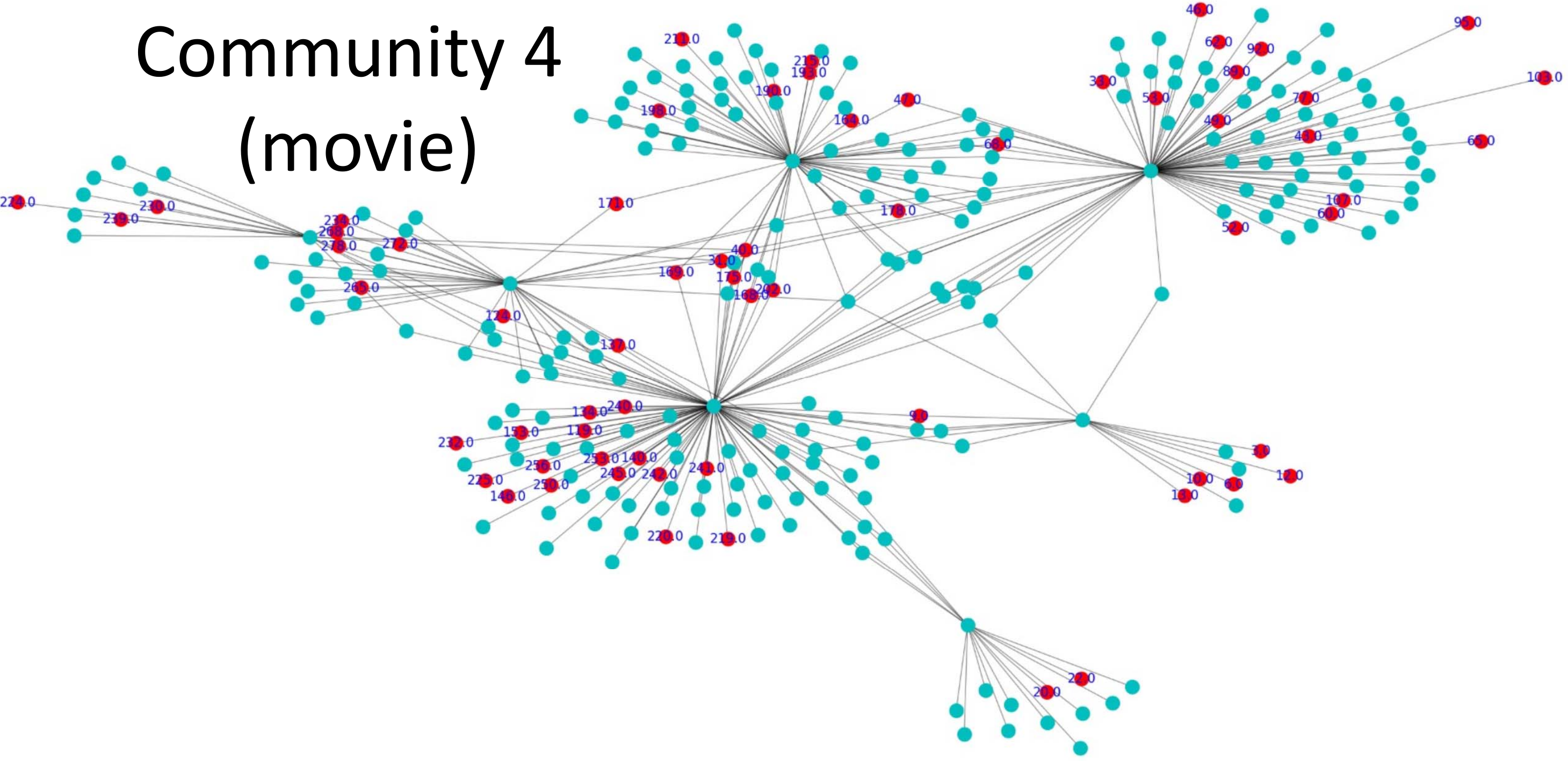}}
  \caption{Recommendation among different communities. The red nodes are the highlighted communities for each result.}
\end{figure}

%

\section{Related works}

Link prediction is an ubiquitous problem in recommendation system \cite{esslimani2011densifying}, social media \cite{10.1007/978-3-319-73198-8_10}, medicine \cite{DBLP:conf/kdd/LichtenwalterLC10} and even finance \cite{10.1007/978-3-319-73198-8_10}. Most of these methods are the discriminative models to classify the unknown links as the existing or the non-existing links for the target networks \cite{menon2011link}. The mainstream methods for the link prediction include the classic methods based on the statistical empirical evidence and the machine learning methods. The classic methods \cite{liben2007link} \cite{adamic2003friends} \cite{0295-5075-61-4-567} measure the similarities for between couples of nodes through their common neighbor numbers \cite{Newman404}. The power law distribution of the complex network \cite{Barab509} helps these similarity scoring methods perform well on distinguishing the positive and negative links statistically. However, since they lack the structure to get the network formation dynamics, they could not capture the accurate links for a real-world network when the statistical rules are not significant enough. To make the accurate prediction for a single node, many machine learning methods apply the matrix factorization \cite{10.1007/978-3-642-23783-6_28} or the graph representation learning methods \cite{Perozzi:2014:DOL:2623330.2623732}. Their resulted latent representations for the users improve the result of the prediction performance on the user-level. Some works further improve the prediction performance by applying a GAN framework to supervise the representing quality \cite{DBLP:conf/aaai/WangWWZZZXG18}. To increase the generality of the graph representation learning methods, recent methods combine the graph convolution and the graph autoencoder (GAE) \cite{DBLP:journals/corr/KipfW16a} to extend the embedding vectors to a higher dimension latent space. Recent work also applies the graph neural network to analyze the link prediction problem \cite{DBLP:conf/nips/ZhangC18}. However, since most mainstream methods ignore the temporal information in the evolving networks, it is difficult for them to accurately distinguish the future positive and negative links with the limit data in a relatively small window. To utilize the temporal information, some work \cite{DBLP:conf/wsdm/YuA017} uses the sliding-window style time-dependent method to address the temporal link prediction issue. Furthermore, several works discuss the related problems as the temporal graph mining \cite{DBLP:conf/kdd/YangYWCZL16} \cite{DBLP:conf/icde/LiSQYD18} by considering the links as streams \cite{DBLP:journals/tcs/ViardLM16}. Whereas, the existing works ignore the important information contained in the chronological order of the link emerging. We orders the links in the sequence with their emerging chronological order to simulate the real scenario and applies the sequence modeling framework \cite{NIPS2014_5346} to capture the temporal link formation patterns. The experiments show that this framework could capture the effective temporal link formation patterns and result in the good performance in predicting the future links based on the observed links.

\section{Conclusion}

In this work, we propose the Generative Link Sequence Modeling (GLSM) to predict the future links based on the historical observation. GLSM combines an RNN process to learn the temporal link formation patterns in the sequence modeling framework and a two-step sampling link generation process to generate the future links. To transform the temporal link prediction to the framework of the sequence modeling, we propose the self-tokenization mechanism to convert the binary link sequence to the unary token sequence with the proper granularity. The self-tokenization process incorporates a clustering process which allows it generates the token sequence automatically. The clustering process also helps the resulted token sequence to capture the network topological information. The RNN process of GLSM learns the temporal link formation pattern from the resulted token sequence. Since the RNN process depends on the obtained token sequence from the self-tokenization process, the RNN and self-tokenization process could be trained simultaneously. With the learned temporal link formation pattern, GLSM generates the future links with the two-step sampling link generation process. Experimental results show that GLSM performs the best of all the mentioned methods on the real-world temporal networks and this verifies the temporal information contained in the chronological order for the links is useful in designing the link prediction models.

\section{Acknowledgments}
This work is done while Yue Wang is visiting the University of Illinois at Chicago. This work is supported by the National Natural Science Foundation of China (Grant No.61503422, 61602535), The
Open Project Program of the National Laboratory of Pattern Recognition (NLPR) and Program for Innovation Research in Central University of Finance and Economics. This work is also supported in part by NSF through grants IIS-1526499, IIS-1763325, and CNS-1626432, and NSFC 61672313.

\bibliographystyle{plain}
\bibliography{ref}

\end{document}